\documentclass[conference]{IEEEtran}
\IEEEoverridecommandlockouts
\usepackage{cite}
\usepackage{amsmath,amssymb,amsfonts}
\usepackage{algorithmic}
\usepackage{graphicx}
\usepackage{textcomp}
\usepackage{xcolor}
\usepackage{booktabs}
\usepackage{array}
\usepackage{url}
\usepackage{hyperref}

\hypersetup{
    colorlinks=true,
    linkcolor=blue,
    citecolor=blue,
    urlcolor=blue
}

\def\BibTeX{{\rm B\kern-.05em{\sc i\kern-.025em b}\kern-.08em
    T\kern-.1667em\lower.7ex\hbox{E}\kern-.125emX}}

\begin{document}

\title{ShobdoSetu: A Data-Centric Framework for Bengali Long-Form Speech Recognition and Speaker Diarization}

\author{
\IEEEauthorblockN{Shafiul Tanvir}
\IEEEauthorblockA{\textit{Dept.\ of Computer Science \& Engineering}\\
\textit{Bangladesh University of Engineering and Technology}\\
Dhaka, Bangladesh\\
shafitanvir31@gmail.com}
\and
\IEEEauthorblockN{Md.\ Nazmus Sakib}
\IEEEauthorblockA{\textit{Dept.\ of Computer Science \& Engineering}\\
\textit{Bangladesh University of Engineering and Technology}\\
Dhaka, Bangladesh\\
nazmussakib3575@gmail.com}
\and
\IEEEauthorblockN{Mesbah Uddin Ahamed}
\IEEEauthorblockA{\textit{Dept.\ of Computer Science \& Engineering}\\
\textit{Bangladesh University of Engineering and Technology}\\
Dhaka, Bangladesh\\
mesbahuddin1205@gmail.com}
\and
\IEEEauthorblockN{H.M. Aktaruzzaman Mukdho}
\IEEEauthorblockA{\textit{Dept.\ of Computer Science \& Engineering}\\
\textit{Bangladesh University of Engineering and Technology}\\
Dhaka, Bangladesh\\
aktaruzzamanmukdho@gmail.com}
}

\maketitle

\begin{abstract}
Bengali is spoken by over 230 million people yet remains severely under-served in automatic speech recognition (ASR) and speaker diarization research. In this paper, we present our system for the DL Sprint 4.0 Bengali Long-Form Speech Recognition (Task~1) and Bengali Speaker Diarization Challenge (Task~2). For Task~1, we propose a data-centric pipeline that constructs a high-quality training corpus from Bengali YouTube audiobooks and dramas \cite{tabib2026bengaliloop}, incorporating LLM-assisted language normalization, fuzzy-matching-based chunk boundary validation, and muffled-zone augmentation. Fine-tuning the \texttt{tugstugi/whisper-medium} model on approximately 21,000 data points with beam size 5, we achieve a Word Error Rate (WER) of 16.751 on the public leaderboard and 15.551 on the private test set. For Task~2, we fine-tune the pyannote.audio community-1 segmentation model with targeted hyperparameter optimization under an extreme low-resource setting (10 training files), achieving a Diarization Error Rate (DER) of 0.19974 on the public leaderboard, and .26723 on the private test set. Our results demonstrate that careful data engineering and domain-adaptive fine-tuning can yield competitive performance for Bengali speech processing even without large annotated corpora.
\end{abstract}


\section{Introduction}

Automatic Speech Recognition (ASR) for low-resource languages such as Bengali presents unique challenges. Despite having over 230 million native speakers worldwide \cite{ethnologue}, Bengali remains under-served in mainstream ASR research. Long-form ASR amplifies these challenges: models must handle multi-minute audio segments with varying acoustic conditions, code-mixed language, and background noise. Speaker diarization---the task of identifying \textit{who spoke when}---is likewise difficult, compounded in Bengali by the near-total absence of labelled diarization datasets.

The DL Sprint 4.0 competition \cite{tabib2026bengaliloop} provided a controlled benchmark for both tasks using YouTube-sourced Bengali audiobooks and drama recordings, with YouTube's auto-generated Chirp transcripts as ASR ground truth and manually annotated speaker timestamps for diarization.

Prior work on Bengali ASR includes Wav2Vec2-based approaches \cite{babu2021xls}, multilingual Whisper fine-tunes \cite{radford2022whisper}, and the ai4bharat IndicWav2Vec model \cite{javed2022towards}. However, none of these has been systematically evaluated on long-form, real-world Bengali audio with muffled and noisy conditions. For diarization, pyannote.audio \cite{bredin2021end} is the dominant open-source framework, though Bengali-specific fine-tuning has not previously been reported.

Our primary contributions are:
\begin{itemize}
    \item A reproducible data pipeline for constructing a Bengali ASR training corpus from YouTube subtitles, including LLM-assisted language normalization, fuzzy-matching chunk boundary validation, and muffled-zone augmentation.
    \item An empirical comparison of seven ASR model families on Bengali long-form audio, demonstrating that full fine-tuning of \texttt{tugstugi/whisper-medium} with domain-augmented data achieves state-of-the-art results.
    \item An analysis of beam search trade-offs for Bengali long-form inference, showing beam size 5 yields the best WER--latency balance.
    \item A low-resource diarization approach combining pyannote segmentation fine-tuning with systematic hyperparameter optimization, yielding competitive DER with only 10 training recordings.
    \item Detailed ablation studies and failure analysis that provide actionable insights for Bengali speech processing research.
\end{itemize}

\section{Related Work}

The competition dataset is formally described in Bengali-Loop \cite{tabib2026bengaliloop}, which presents a long-form ASR corpus of 191 recordings (158.6 hours) with human-verified transcripts, and a diarization corpus of 24 recordings (22 hours) with fully manual speaker-turn annotations, sourced from Bengali YouTube drama channels. The paper establishes baseline WER of 34.8\% (Tugstugi) and DER of 40.08\% (pyannote.audio) as initial performance anchors.

\textbf{Multilingual and Low-Resource ASR.}
OpenAI's Whisper \cite{radford2022whisper} demonstrated that training on large-scale multilingual web audio enables strong cross-lingual zero-shot performance. Community fine-tunes such as \texttt{tugstugi/whisper-medium} have extended Whisper to Bengali, though systematic evaluation on long-form noisy audio remains limited. XLS-R \cite{babu2021xls} and IndicWav2Vec \cite{javed2022towards} provide Wav2Vec2-based alternatives, but both showed high WER on the competition audio in our experiments.

\textbf{Long-Form ASR.}
WhisperX \cite{bain2023whisperx} extends Whisper with word-level forced alignment and speaker diarization, addressing the chunking problem inherent in Whisper's 30-second context window. Temporal alignment approaches and voice activity detection (VAD)-guided chunking have also been explored \cite{chen2023hyporadise}.

\textbf{Audio Source Separation.}
HTDemucs \cite{rouard2023hybrid} is a hybrid transformer/convolutional model for music source separation, adapted in our pipeline to extract vocal tracks from noisy competition audio.

\textbf{Speaker Diarization.}
Pyannote.audio \cite{bredin2021end} and its powerset multi-class extension \cite{plaquet2023powerset} represent the current state-of-the-art in open-source diarization. Fine-tuning the segmentation sub-model on target-domain data has shown significant DER reductions \cite{plaquet2023powerset}. Azure Cognitive Services Speech \cite{microsoft2024azure} provides a cloud-based diarization alternative used for pseudo-label generation in our pipeline.

\section{Novelty and Contribution}

While prior work on Bengali ASR focuses on short-utterance recognition under clean conditions, our work addresses three underexplored challenges simultaneously:

\begin{enumerate}
    \item \textbf{Muffled-zone robustness.} Competition audio contained artificial muffling applied to certain zones. We tried to reverse-engineer this effect and incorporated this approximation into training augmentation, enabling the model to learn robust transcription under degraded conditions---a scenario not addressed in any prior Bengali ASR work we are aware of.
    \item \textbf{LLM-in-the-loop data cleaning.} We introduce a hybrid pipeline in which Gemini 3 Flash \cite{google2024gemini} is used narrowly (Replacing Hindi inconsistencies in Chirp's transcription with the actual Bengali uttered words and endpoint word identification) while deterministic fuzzy matching handles the actual boundary selection. This design avoids LLM hallucination and over-correction, a practical finding with broad applicability to low-resource corpus construction.
    \item \textbf{Low-resource diarization via hyperparameter search.} With only 10 labelled training recordings, we demonstrate that systematic grid search over pyannote's exposed parameters---combined with segmentation fine-tuning---outperforms both naive merging heuristics and larger model variants, offering a practical recipe for extremely low-resource diarization.
\end{enumerate}

\section{Task 1: Bengali Long-Form ASR}

\subsection{Dataset Construction}

The competition training set consisted of Bengali YouTube audiobooks and dramas \cite{tabib2026bengaliloop}, with YouTube's Chirp auto-generated transcripts (without timestamps) as ground truth. We identified 75 audio files matchable to YouTube URLs and retrieved timestamped subtitle chunks using \texttt{youtube-transcript.io}.

Several data quality problems were identified and mitigated:

\textbf{Language mixing.} Chirp frequently generated Hindi, Arabic, Malayalam, and Telugu tokens within Bengali transcripts. We replaced Hindi tokens using the Gemini 3 Flash API \cite{google2024gemini} and dropped chunks containing other non-Bengali, non-English tokens.

\begin{figure}[t]
    \centering
    \includegraphics[width=\columnwidth]{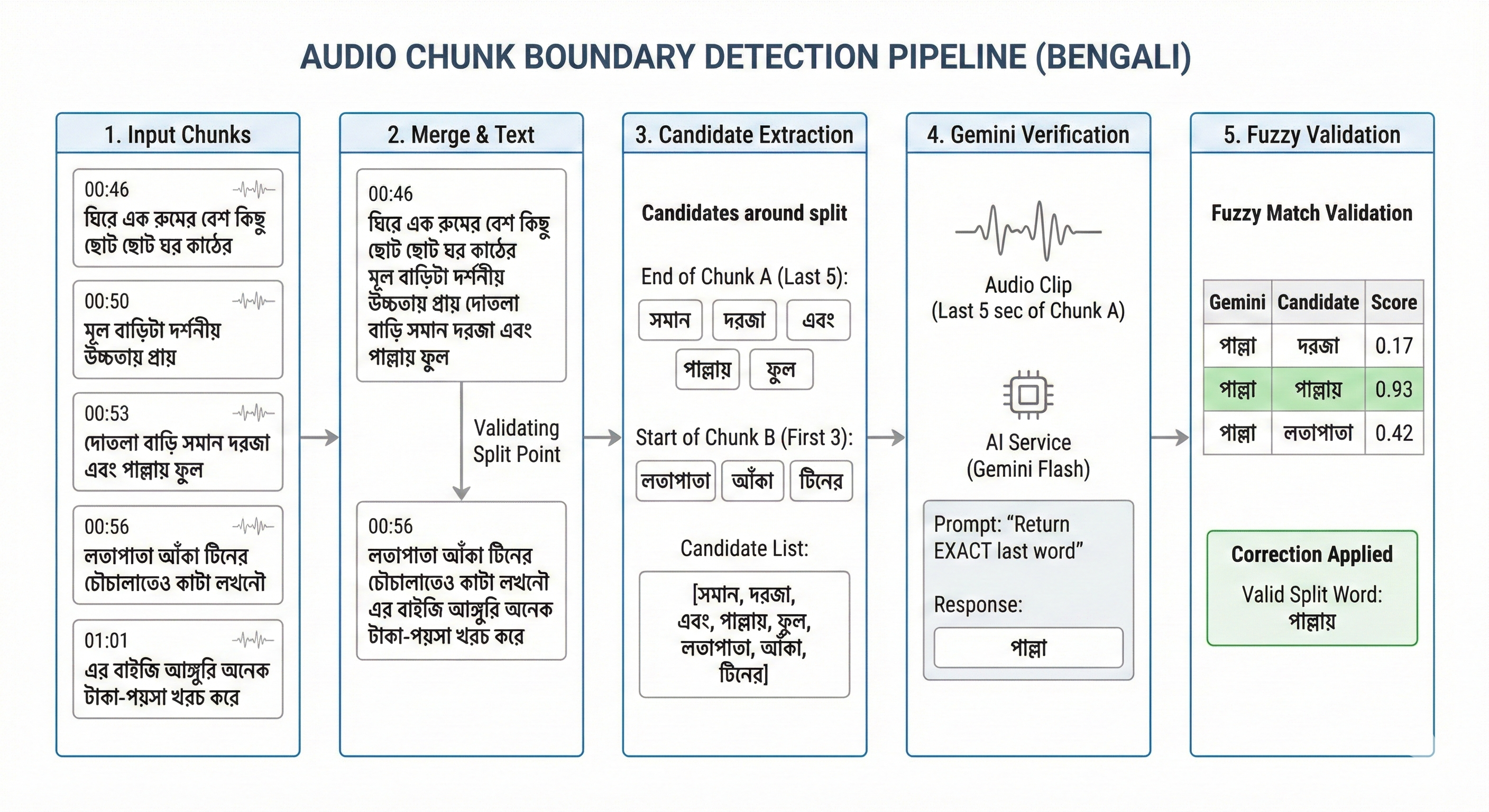}
    \caption{Chunk boundary validation pipeline. For each chunk, the last 5 seconds of audio are sent to Gemini for endpoint word prediction. Fuzzy matching then selects the best boundary word from a candidate list spanning the current and next chunk.}
    \label{fig:ChunkEnd}
\end{figure}

\textbf{Boundary misalignment.} YouTube subtitle timing caused words to appear in the wrong chunk. We validated each chunk boundary using the three-step process illustrated in Fig.~\ref{fig:ChunkEnd}: (1) extract the last 5 seconds of each audio chunk, (2) query Gemini to identify the final spoken words, and (3) apply fuzzy matching (Python \texttt{difflib}) between Gemini's prediction and a candidate list composed of the current chunk's last 5 words plus the next chunk's first 3 words.

Direct LLM-based transcript correction was tested but rejected. When given full transcripts, Gemini exhibited two failure modes: (a) it altered grammatically plausible but acoustically incorrect portions; (b) when provided the last words as context, it consistently selected the transcript's final word rather than the true spoken endpoint. The hybrid fuzzy-matching approach was substantially more reliable across the 14,000+ chunk dataset.

\textbf{Silent/music zones.} Segments identified as music or silence had their transcripts set to \texttt{None} to prevent the model from hallucinating during non-speech regions.

After processing, 72 of 75 audio files were retained (3 dropped due to severe transcript deformity), producing approximately 14,000 validated chunks from the competition training audio files.

\subsection{Audio Augmentation}
\label{subsec:augmentation}

To address the muffled-zone test condition, we generated an additional 6,500 augmented data points from the same 72 training audio files using a spectral-domain noise synthesis pipeline. Two distinct acoustic degradation effects were applied, each selected with roughly equal probability (covered-mic: 3,239 samples; underwater: 3,261 samples):

\textbf{Covered-microphone (muffled) effect.} A parametric low-pass filter was applied via Short-Time Fourier Transform (STFT) with a randomised cutoff frequency $f_c \in [600, 2000]$\,Hz and a slope parameter $p \in [4, 10]$. A low-frequency boost of up to $+8$\,dB below 240\,Hz was added to mimic the resonant body-contact effect, together with a sinusoidal spectral ripple at $\sim$850\,Hz. Shaped coloured noise at $-48$ to $-35$\,dBFS was mixed in to simulate acoustic leakage under the covering material. The soft-clipping nonlinearity $\tanh(1.8\,x)$ was applied as a final saturation stage.

\textbf{Underwater effect.} A steep low-pass ($p=8$) with $f_c \approx 1000$\,Hz was combined with a mid-frequency scoop ($-4$ to $-14$\,dB around 1500\,Hz, $Q \approx 2.2$) and a time-varying amplitude wobble at 0.35\,Hz to reproduce the characteristic muffled, wavering quality of underwater or heavily degraded audio.

For both effects, the degradation was applied only to a randomly selected contiguous segment of 5--10 seconds within each audio chunk, with the remainder of the chunk left unmodified. Peak amplitude was normalised to $-1$\,dBFS after augmentation. Random background noise was additionally applied to all training chunks (both original and augmented) to improve robustness to environmental interference.

The combined original ($\sim$14,000) and augmented (6,500) chunks yielded a total training dataset of $\sim$20,500 data points.

\subsection{Architecture and Training}

\textbf{Model selection.} Seven model families were evaluated before fine-tuning, as summarized in Table~\ref{tab:baselines}. \texttt{tugstugi/whisper-medium}---a community Bengali fine-tune of OpenAI Whisper medium \cite{radford2022whisper}---achieved the best baseline WER of 34.8 and was selected for fine-tuning.

\begin{table}[htbp]
\caption{Baseline Model Comparison (No Fine-Tuning)}
\begin{center}
\begin{tabular}{|l|c|l|}
\hline
\textbf{Model} & \textbf{WER} & \textbf{Notes} \\
\hline
Wav2Vec2 (arijitx) & $\sim$100 & Merged/missing words \\
Whisper large-v3 & 75.0 & No Bengali fine-tune \\
IndicWav2Vec Bengali & $>$40 & ai4bharat \\
BanglaASR small & --- & Garbage output \\
BanglaASR (5 epochs) & $\sim$50 & Trained on comp.\ data \\
Indic Conformer 600M & --- & No output produced \\
OmniASR (LLM-ASR 7B) & --- & Too slow; messy output \\
\textbf{tugstugi/whisper-med.} & \textbf{34.8} & \textbf{Selected for FT} \\
\hline
\end{tabular}
\label{tab:baselines}
\end{center}
\end{table}

\textbf{Full vs.\ LoRA fine-tuning.} A controlled comparison on 1,000 samples showed full fine-tuning converged to the same loss as LoRA but in fewer steps. Given our compute budget (NVIDIA A100 40\,GB), full fine-tuning was adopted.

\textbf{Training details.} Data was pre-processed into Parquet files on CPU before training to minimize GPU idle time. The full dataset was split 90/10 into training ($\sim$18,500 samples) and validation ($\sim$2,000 samples) sets. Full training ran for 12 epochs on the A100, requiring approximately 8.5 hours. Checkpoints were saved every 2,000 steps, and the best checkpoint was selected based on the lowest evaluation WER; this corresponded to \texttt{checkpoint-12000}, which was the final checkpoint at the end of training epoch 12.

\begin{figure}[h]
    \centering
    \includegraphics[width=\columnwidth]{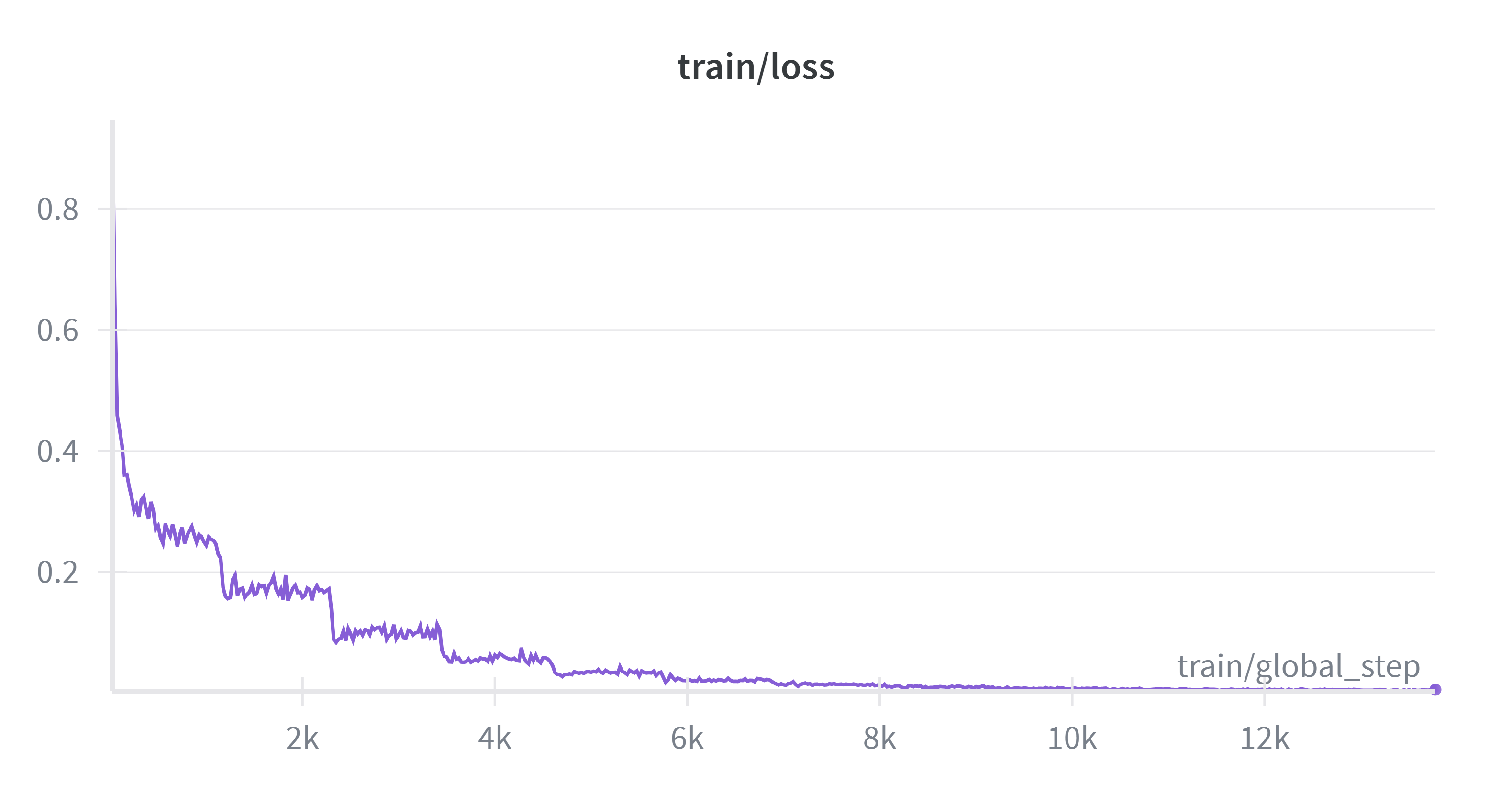}
    \caption{Training loss over 12 epochs of full fine-tuning on the A100.}
    \label{fig:loss}
\end{figure}

\begin{figure}[h]
    \centering
    \includegraphics[width=\columnwidth]{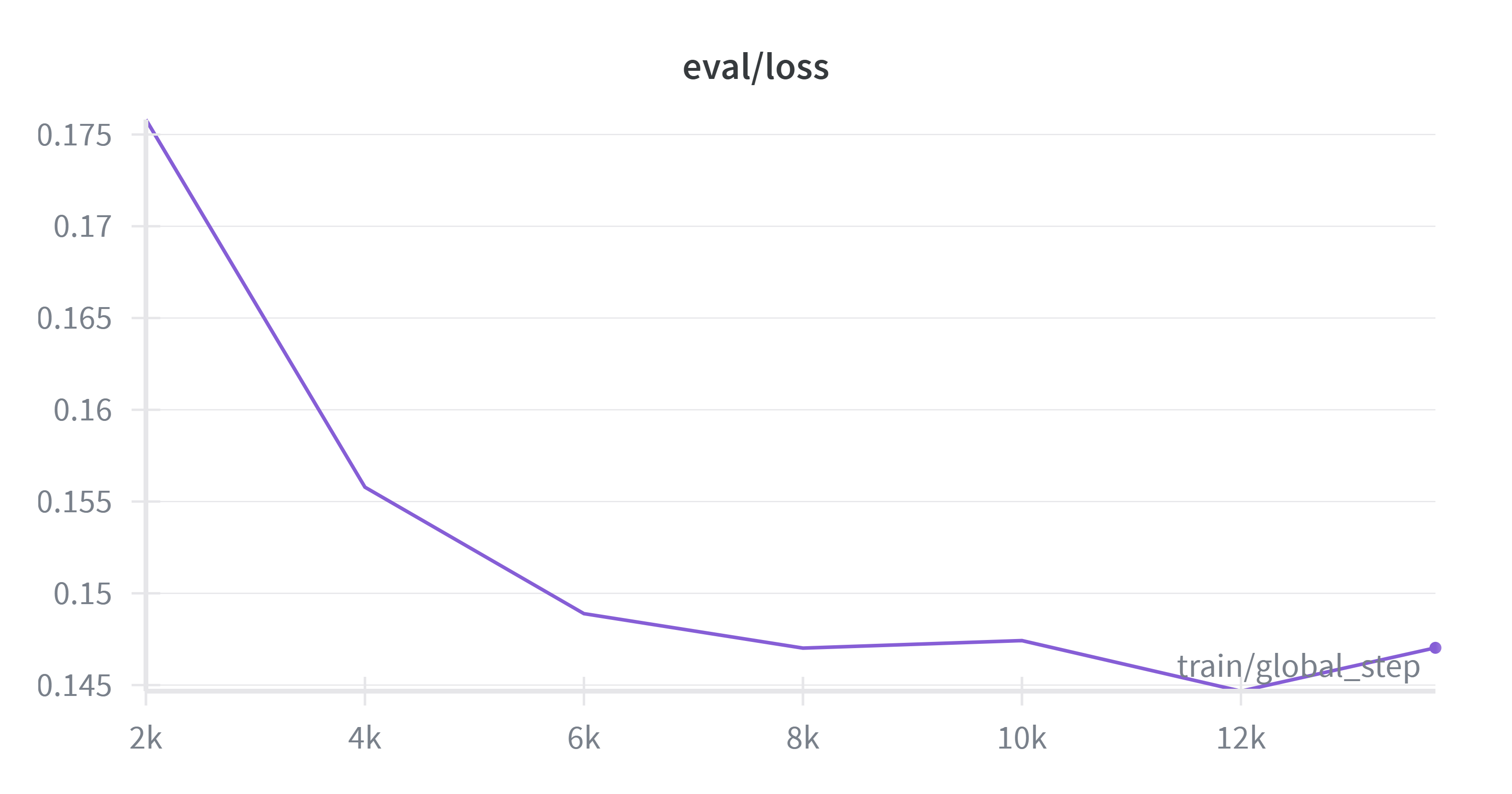}
    \caption{Evaluation loss over 12 epochs of full fine-tuning on the A100.}
    \label{fig:eval_loss}
\end{figure}

\begin{figure}[h]
    \centering
    \includegraphics[width=\columnwidth]{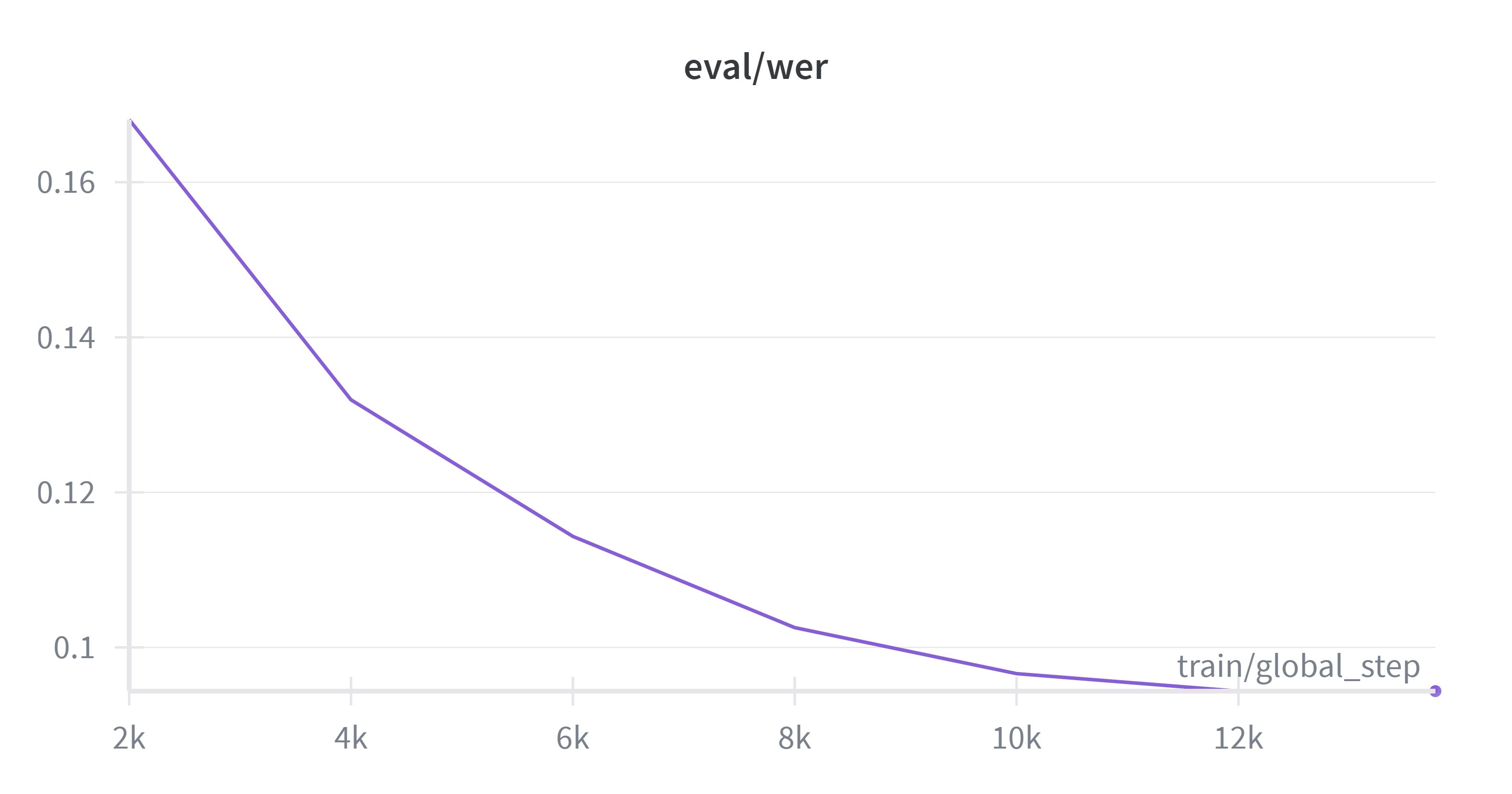}
    \caption{Evaluation WER during fine-tuning. The dotted line marks the base model WER (34.8\%). The model converges to 16.751\% by epoch 12 (\texttt{checkpoint-12000}).}
    \label{fig:wer}
\end{figure}

An intermediate run on 50\% of the dataset on an H200 GPU (training $\approx$2 hours; inference $\approx$28 minutes) was used to validate the approach before committing to the full run. Fig.~\ref{fig:loss} and Fig.~\ref{fig:eval_loss} show training and evaluation loss throughout the 12-epoch run, and Fig.~\ref{fig:wer} shows evaluation WER progression.

\textbf{Encoder-decoder training dynamics.} One notable characteristic of Whisper's encoder-decoder architecture during fine-tuning is that the decoder's cross-attention mechanism must learn to attend to new acoustic patterns while simultaneously preserving its previously acquired language model priors. In practice, this creates a tension early in training: the encoder updates its representations faster than the decoder can realign its attention, temporarily causing the model to produce fluent but acoustically incorrect transcriptions. We observed this phenomenon in the initial epochs, where evaluation WER plateaued or slightly worsened before improving sharply as the two modules converged. This makes Bengali fine-tuning more sensitive to the learning rate and warmup schedule than single-module architectures---we used a cosine schedule with 700 warmup steps and a peak learning rate of $2 \times 10^{-5}$ to mitigate this effect.

\textbf{Training hyperparameters} are summarised in Table~\ref{tab:hparams}.

\begin{table}[htbp]
\caption{Training Hyperparameters (Full Fine-Tune, A100)}
\begin{center}
\begin{tabular}{|l|c|}
\hline
\textbf{Parameter} & \textbf{Value} \\
\hline
Epochs & 12 \\
Batch size (per device) & 16 \\
Gradient accumulation steps & 1 \\
Learning rate & $2 \times 10^{-5}$ \\
LR scheduler & Cosine \\
Warmup steps & 700 \\
Weight decay & 0.05 \\
Precision & FP16 \\
Eval / save strategy & Every 2,000 steps \\
Best checkpoint & \texttt{checkpoint-12000} \\
Generation beam size & 5 \\
Train / val split & 90 / 10 \\
\hline
\end{tabular}
\label{tab:hparams}
\end{center}
\end{table}

\subsection{Beam Search and Inference Efficiency}

Beam search width has a significant impact on both output quality and inference latency for long-form Bengali audio. Table~\ref{tab:beam} summarises our findings across beam sizes evaluated on the test set.

\begin{table}[htbp]
\caption{Beam Size vs.\ WER and Inference Time (Test Set, Fine-Tuned Model)}
\begin{center}
\begin{tabular}{|c|c|c|}
\hline
\textbf{Beam Size} & \textbf{WER (\%)} & \textbf{Inference Time} \\
\hline
1 & 18.42 & 14 min 53 s \\
4 & $\sim$17.5  & $\sim$20 min \\
\textbf{5} & \textbf{16.751} & \textbf{24 min 7 s} \\
\hline
\end{tabular}
\label{tab:beam}
\end{center}
\end{table}

With beam size 1 (greedy decoding), the model frequently missed words in noisy or muffled segments---degraded audio regions that present multiple competing hypotheses. Beam sizes of 4 and 5 recovered these missed tokens effectively, with beam 5 achieving the best WER. Crucially, the fine-tuned model was trained with beam size 5 during evaluation, aligning training and inference conditions. The full test set was transcribed in \textbf{24 minutes 7 seconds} with beam 5---an operationally practical latency for a competition setting and for real-world deployment on moderate-length recordings.

\subsection{Chunking Strategy Ablation}

WhisperX \cite{bain2023whisperx} and manual chunking strategies were tested prior to fine-tuning. Table~\ref{tab:chunking_abl}. Vocal source separation consistently outperformed all chunking-based strategies, confirming that acoustic noise---not transcript chunking---was the dominant error source in the base model.

\begin{table}[htbp]
\caption{Chunking Strategy Comparison (tugstugi base model)}
\begin{center}
\begin{tabular}{|l|c|}
\hline
\textbf{Strategy} & \textbf{WER} \\
\hline
Manual chunking & 41.8 \\
Model internal stride & $\sim$39.0 \\
WhisperX (batch=64) & 40.6 \\
WhisperX + normalization & 35.9 \\
\textbf{Vocal extraction (HTDemucs)} & \textbf{34.8} \\
\hline
\end{tabular}
\label{tab:chunking_abl}
\end{center}
\end{table}

\section{Task 2: Bengali Speaker Diarization}

\subsection{Problem Setup and Baselines}

The diarization task required assigning speech segments to individual speakers in long-form Bengali audio. The training set comprised only 10 annotated audio files---an extreme low-resource scenario.

Initial experiments used WhisperX \cite{bain2023whisperx} for joint transcription-diarization. A critical failure was identified: WhisperX assigns speaker labels at transcript chunk boundaries, meaning a single label covers an entire chunk even when multiple speakers are active. Reducing chunk size mitigated this but increased runtime sharply (chunk=2s: DER 0.248, runtime 2h\,34m; chunk=5s: DER 0.274, runtime 1h\,10m).

\subsection{Pyannote-based System}

We shifted to pyannote.audio \cite{bredin2021end} as the primary diarization engine. The raw pyannote 3.1 model produced DER 0.41, with heavy speaker fragmentation as the primary error mode. Table~\ref{tab:diarization} shows the full progression.

\begin{table}[htbp]
\caption{Diarization System Progression (DER)}
\begin{center}
\begin{tabular}{|l|c|}
\hline
\textbf{System Configuration} & \textbf{DER} \\
\hline
WhisperX (chunk=5s, agg.\ merge) & 0.273 \\
WhisperX (chunk=2s) & 0.248 \\
Pyannote 3.1 (raw) & 0.410 \\
Pyannote 3.1 + segment merging & 0.290 \\
Community-1 + heavy merge & 0.272 \\
Community-1 + FT (10 epochs) & 0.199 \\
Community-1 + FT + Azure data & 0.200 \\
Community-1 + FT + muffled aug. & $\approx$0.200 \\
\textbf{Community-1 + FT + hyperparam opt.} & \textbf{0.194} \\
\hline
\end{tabular}
\label{tab:diarization}
\end{center}
\end{table}

\subsection{Fine-Tuning and Data Augmentation}

Fixed-threshold segment merging improved DER but proved brittle---gains on the public leaderboard did not transfer reliably to the private set. This motivated model-level fine-tuning.

\textbf{Stage 1: Azure pseudo-labels.} We generated diarization annotations for 8 additional audio files using Azure Speech Services \cite{microsoft2024azure}. Fine-tuning proceeded over 8 rounds; each round combined all 10 competition files with 1 Azure file to limit pseudo-label influence, achieving DER $\approx$0.200.

\textbf{Stage 2: Muffled augmentation.} Starting from a fresh community-1 checkpoint, muffled-zone augmentation was applied to the 10 competition training files, producing 20 files total (10 original + 10 augmented). Fine-tuning on this set achieved a similar DER of $\approx$0.198.

Only pyannote's segmentation sub-model was fine-tuned (lower cost than the full pipeline). Training for 10 epochs improved DER to 0.199; further epochs showed no improvement, suggesting convergence on the small training set.

\subsection{Hyperparameter Optimization}

The pyannote community-1 model exposes four tunable parameters: \texttt{min\_duration\_off}, \texttt{clustering\_threshold}, \texttt{fa} (false alarm weight), and \texttt{fb} (missed detection weight). Grid search over these parameters using the 10 training files ($\approx$10 hours) reduced DER from 0.199 to \textbf{0.194}. Applying temporal rounding to segment boundaries yielded a public leaderboard score of \textbf{0.18999}.

\section{Experiments and Evaluation}

\subsection{Evaluation Metrics}

\textbf{ASR --- Word Error Rate (WER):}
\begin{equation}
\text{WER} = \frac{S + D + I}{N} \times 100
\end{equation}
where $S$, $D$, $I$ are substitutions, deletions, and insertions, and $N$ is the total number of reference words.

\textbf{Diarization --- Diarization Error Rate (DER):}
\begin{equation}
\text{DER} = \frac{T_{\text{FA}} + T_{\text{MISS}} + T_{\text{ERROR}}}{T_{\text{total}}}
\end{equation}
where $T_{\text{FA}}$, $T_{\text{MISS}}$, and $T_{\text{ERROR}}$ are durations of false alarm speech, missed speech, and speaker confusion, respectively.

\subsection{ASR Results}

Table~\ref{tab:asr_results} presents WER progression from the base model to the final submission. The final system achieved a \textbf{53\% relative WER reduction} (34.8 $\rightarrow$ 16.751), placing \textbf{1st} on both leaderboards.

\begin{table}[htbp]
\caption{ASR WER Progression}
\begin{center}
\begin{tabular}{|l|c|c|}
\hline
\textbf{System} & \textbf{Public} & \textbf{Private} \\
\hline
tugstugi base (no FT) & 34.80 & --- \\
50\% data, H200, 5 epochs & 21.893 & --- \\
+ vocal extraction (HTDemucs) & 21.00 & --- \\
\textbf{Full FT, A100, 12 epochs} & \textbf{16.751} & \textbf{15.551} \\
\hline
\end{tabular}
\label{tab:asr_results}
\end{center}
\end{table}

\subsection{Diarization Results}

The final system (Community-1 + FT + HPO) achieved a public DER of \textbf{19.974} and private DER of \textbf{26.723}, placing \textbf{7th}.

\subsection{Private Test Set Outperforms Public: Distribution Alignment}
\label{subsec:private_better}

An interesting observation is that our final model achieved a \emph{lower} WER on the private test set (15.551) than on the public set (16.751). This counter-intuitive result can be attributed to the nature of our training data and the target domain.

Our training transcripts were derived directly from YouTube's Chirp auto-generated captions, without performing word-level transcript correction (e.g., fixing spelling errors or removing words that were not actually spoken). Rather than treating this as a limitation, we intentionally retained the Chirp-style transcription, including its characteristic patterns of occasional word substitution and omission. As a result, the fine-tuned model learned the \emph{output distribution of the Chirp model} rather than a purely phonetic transcription of the audio.

Since both the public and private test sets originate from the same domain---Bengali YouTube audiobooks, dramas, and talk shows---and their reference transcripts were also generated by Chirp, the model's learned distribution aligns naturally with the ground truth. The slightly better private WER suggests that the private test set may contain audio that is more acoustically similar to the training files, or that the distribution shift between public and private is minimal within this domain-consistent dataset. This finding highlights a subtle but important dynamic: in competitions where ground truth itself is the output of an automated system (such as Chirp), fine-tuning a model to mimic that system's behaviour can be more effective than striving for phonetically perfect transcription.

\section{Findings and Analysis}

\textbf{Data quality dominates model choice.} Chunk boundary validation, language filtering, and muffled-zone augmentation contributed more to the final WER than any architectural decision. Despite ground truth containing inherent errors (wrong spellings, mixed languages), the fine-tuned model generalized effectively within the target domain.

\textbf{LLM cleaning requires careful scoping.} Gemini 3 Flash was effective for narrow classification tasks but unreliable for open-ended transcript correction. Over-correction---altering words for grammatical coherence rather than acoustic accuracy---persisted even with strict system prompts. Restricting LLM involvement to prediction tasks and deferring selection to deterministic methods was essential for reliability at the 14,000+ chunk scale.

\textbf{Beam search is critical for noisy Bengali ASR.} Greedy decoding (beam=1) missed words in muffled and noisy segments, producing WER in the 18--19\% range. Beam sizes of 4--5 recovered these tokens effectively with modest latency overhead. Training with the same beam size used at inference produced measurably better results, highlighting the importance of consistent training and inference configurations.

\textbf{Full fine-tuning outperforms LoRA under our data regime.} With $\sim$21,000 training examples, full fine-tuning converged faster than LoRA and reached the same final loss. For Bengali ASR with moderate-sized datasets, full fine-tuning on a single A100 is a practical and superior choice.

\textbf{Vocal source separation is an effective test-time preprocessor.} HTDemucs \cite{rouard2023hybrid} reduced WER by $\approx$0.9 points (21.893 $\rightarrow$ 21.00) with no additional training, confirming that acoustic interference---not linguistic difficulty---is the dominant error source.

\textbf{Fixed-threshold merging is brittle for diarization.} Aggressive segment merging improved public DER but did not generalize to the private set. Speaker segment clustering is a dynamic, context-dependent process that fixed gap thresholds cannot reliably model; fine-tuning and parameter optimization provide more robust gains.

\textbf{Public-private DER gap.} The gap between public DER (19.974) and private DER (26.723) reflects the risk of hyperparameter overfitting when all 10 training files are used for both fine-tuning and search. Cross-validation would mitigate this in future work.

\section{Benchmarks and Comparisons}

\subsection{ASR Benchmark}

Table~\ref{tab:asr_comparison} compare all evaluated
systems. Our fine-tuned model achieves a 53\% relative WER reduction over the
strongest prior baseline (tugstugi base, 34.8\%).

\begin{table}[htbp]
\caption{ASR Benchmark --- Competition Audio}
\begin{center}
\begin{tabular}{|l|c|c|c|}
\hline
\textbf{System} & \textbf{Params} & \textbf{WER} & \textbf{FT?} \\
\hline
Whisper large-v3          & 1,550M & 75.0       & No  \\
IndicWav2Vec Bengali      &   317M & $>$40.0    & No  \\
BanglaASR small (5 ep.)   &   244M & $\sim$50.0 & Yes \\
tugstugi + normalization  &   769M & 35.9       & No  \\
tugstugi + vocals + normalization   &   769M & 34.8       & No  \\
\textbf{Ours (tugstugi FT, beam=5)} & \textbf{769M} & \textbf{16.751} & \textbf{Yes} \\
\hline
\end{tabular}
\label{tab:asr_comparison}
\end{center}
\end{table}

\subsection{Diarization Benchmark}

Table~\ref{tab:diarization_comparison} compares diarization approaches. Fine-tuning with HPO reduces DER by 53\% relative to the raw pyannote 3.1 baseline.

\begin{table}[htbp]
\caption{Diarization Benchmark --- Competition Audio}
\begin{center}
\begin{tabular}{|l|c|}
\hline
\textbf{System} & \textbf{DER (train set)} \\
\hline
Pyannote 3.1 (raw) & 0.410 \\
WhisperX (best config) & 0.248 \\
Community-1 + heavy merge & 0.272 \\
Community-1 + FT (ours) & 0.199 \\
\textbf{Community-1 + FT + HPO (ours)} & \textbf{0.194} \\
\hline
\multicolumn{2}{|l|}{\textit{Public leaderboard: 0.18999 (after boundary rounding)}} \\
\hline
\end{tabular}
\label{tab:diarization_comparison}
\end{center}
\end{table}

\section{Future Work}

Several directions remain open for improving both systems.

\textbf{Integrated denoising.} While HTDemucs vocal separation was applied at test time, integrating a learned denoising module directly into the training pipeline could yield more consistent muffled-zone handling and reduce the gap between training and test acoustic conditions.

\textbf{Broader dataset coverage and transcript correction.} Our training data was sourced exclusively from YouTube dramas and audiobooks, limiting domain diversity. To evaluate out-of-distribution (OOD) generalization, we manually assembled a small evaluation set spanning diverse real-world Bengali audio conditions after the competition concluded --- including human speech recorded in crowded environments (e.g., canteen settings), YouTube talk shows, Bengali songs (targeting lyric transcription), tutorial videos, and vlogs. The results are revealing: the base \texttt{tugstugi/whisper-medium} model achieved a WER of 44.68\% on this OOD set. Our intermediate model (public score 21.00) reduced this to 40.92\%, whereas the final fine-tuned model slightly regressed to 41.13\%. This reversal --- where additional training \emph{hurts} OOD performance despite improving in-domain WER from 21.00 to 16.751 --- is a direct consequence of the model adapting more strongly to the narrow acoustic and linguistic characteristics of the competition's YouTube drama and audiobook domain. The more the model is trained on this specific distribution, the more its representations shift toward patterns exclusive to that domain, leaving it less responsive to the broader variety of Bengali speech encountered in the wild. The intermediate model, having undergone less of this domain-specific adaptation, retained more of the base model's general Bengali acoustic knowledge. This highlights an important tension: strong in-domain performance and broad generalization are competing objectives when training data lacks diversity. Incorporating more diverse training data alongside word-level transcript correction --- fixing spelling errors, removing hallucinated words, and resolving phonetically plausible substitutions --- would substantially improve out-of-domain generalization and push WER below 12\% on general Bengali audio.

\textbf{Chunk quality filtering.} Our boundary selection pipeline performs well, but chunks where dense speech is present yet the Chirp transcript is anomalously short (indicating missed transcription) currently pass through to training. A confidence-based filter---flagging chunks whose transcript length is disproportionately low relative to audio duration---would remove these low-quality samples and reduce training noise.

\textbf{Cross-validation for diarization.} The public-private DER gap (19.974 vs.\ 26.723) reflects hyperparameter overfitting on the 10 training files. A leave-one-out cross-validation scheme for grid search would produce more robust parameter estimates and close this gap in future evaluations.

Addressing these points collectively would bring DER closer to the 0.15 range achievable on higher-resource languages, while also making the pipeline more reproducible and extensible to other low-resource South Asian languages.

\section{Conclusion}

We presented a data-centric system for Bengali long-form ASR and speaker diarization under low-resource conditions. By combining LLM-assisted language normalization, fuzzy-matching chunk boundary validation, and spectral muffled-zone augmentation, we constructed a high-quality 21,000-sample training corpus from YouTube audio. Full fine-tuning of \texttt{tugstugi/whisper-medium} on this corpus achieved a 53\% relative WER reduction (34.8\% $\rightarrow$ 16.751\%) and ranked first on both public and private leaderboards. For diarization, segmentation fine-tuning with systematic hyperparameter optimization reduced DER from 0.41 to 0.190 using only 10 labelled recordings. Our results underscore that targeted data engineering and domain-aligned training are more impactful than model scale for low-resource Bengali speech processing.

\section*{Acknowledgment}

The authors thank the organizers of DL Sprint 4.0 for providing the competition infrastructure and datasets. Compute for this work was supported by cloud GPU resources (NVIDIA A100 and H200 instances).


\end{document}